%% file: arXiv.tex
\documentclass[letterpaper]{article} 
\usepackage[submission]{aaai25}  
\usepackage{times}  
\usepackage{helvet}  
\usepackage{courier}  
\usepackage[hyphens]{url}  
\usepackage{graphicx} 
\usepackage{natbib}  
\usepackage{caption} 
\usepackage{algorithm}
\usepackage{algorithmic}

\usepackage{xcolor}
\usepackage{babel}
\usepackage{textcomp}
\usepackage{multirow}
\usepackage{amstext}
\usepackage{subfig}
\usepackage{booktabs}

\usepackage{newfloat}
\usepackage{listings}
\DeclareCaptionStyle{ruled}{labelfont=normalfont,labelsep=colon,strut=off} 
\floatstyle{ruled}
\newfloat{listing}{tb}{lst}{}
\floatname{listing}{Listing}
\title{CapHDR2IR: Caption-Driven Transfer from Visible Light to Infrared Domain}

\author {
    Jingchao Peng\textsuperscript{\rm 1}\textsuperscript{\rm 2}, 
    Thomas Bashford-Rogers\textsuperscript{\rm 1}, 
    Zhuang Shao\textsuperscript{\rm 3}, 
    Haitao Zhao\textsuperscript{\rm 2} \thanks{Corresponding author.}, 
    Aru Ranjan Singh\textsuperscript{\rm 1}, 
    Abhishek Goswami\textsuperscript{\rm 1}, 
    Kurt Debattista\textsuperscript{\rm 1}
}
\affiliations {
    \textsuperscript{\rm 1}WMG, University of Warwick\\
    \textsuperscript{\rm 2}School of Information Science and Technology, East China University of Science and Technology\\
    \textsuperscript{\rm 3}School of Engineering, Newcastle University\\
}

\begin{document}

\twocolumn[{
\renewcommand\twocolumn[1][]{#1}
\maketitle
\vspace{-0.5cm}
}]

\begin{abstract}
Infrared (IR) imaging offers advantages in several fields due to its unique ability of capturing content in extreme light conditions. However, the demanding hardware requirements of high-resolution IR sensors limit its widespread application. As an alternative, visible light can be used to synthesize IR images but this causes a loss of fidelity in image details and introduces inconsistencies due to lack of contextual awareness of the scene. This stems from a combination of using visible light with a standard dynamic range, especially under extreme lighting, and a lack of contextual awareness can result in pseudo-thermal-crossover artifacts. This occurs when multiple objects with similar temperatures appear indistinguishable in the training data, further exacerbating the loss of fidelity. To solve this challenge, this paper proposes CapHDR2IR, a novel framework incorporating vision-language models using high dynamic range (HDR) images as inputs to generate IR images. HDR images capture a wider range of luminance variations, ensuring reliable IR image generation in different light conditions. Additionally, a dense caption branch integrates semantic understanding, resulting in more meaningful and discernible IR outputs. Extensive experiments on the HDRT dataset show that the proposed CapHDR2IR achieves state-of-the-art performance compared with existing general domain transfer methods and those tailored for visible-to-infrared image translation.

\end{abstract}

\input{Sec1_Introduction}

\input{Sec2_Related_Works}

\input{Sec3_Methodology}

\input{Sec4_Experiments}

\input{Sec5_Conclusion}

\bibliography{aaai25}

\end{document}

%% file: Sec1_Introduction.tex
\section{Introduction}

\begin{figure}[!t]
\subfloat[\label{fig:Comparison-between-SDR}HDR images provide more content information than SDR.]{
    \hspace*{-0.05\columnwidth}
    \includegraphics[width=1.05\columnwidth]{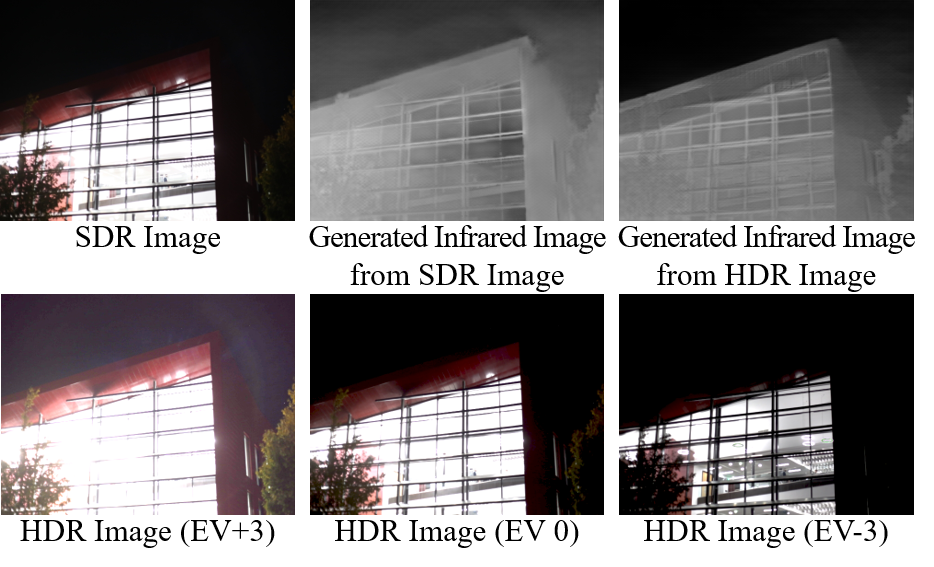}
    \vspace{-0.2cm}
}
\par\vspace{-0.1cm}
\subfloat[\label{fig:Vision-language-model-is}The vision-language model can bridge the gap between visible light and infrared, enhancing scene context understanding.]{
    \hspace*{-0.05\columnwidth}
    \includegraphics[width=1.05\columnwidth]{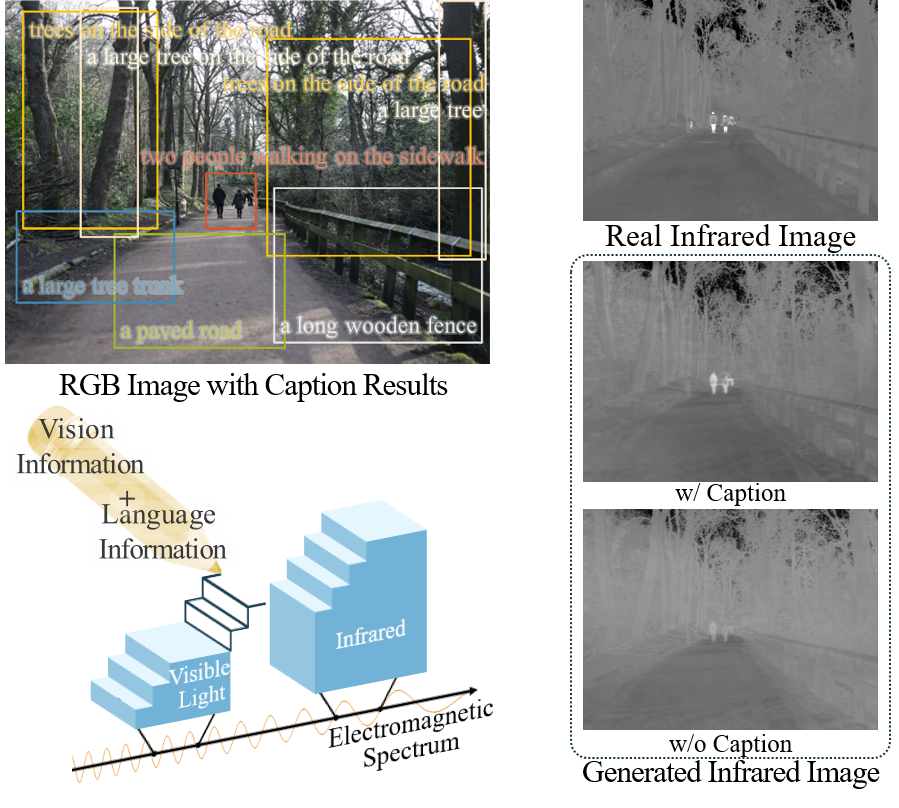}
    \vspace{-0.2cm}
}\vspace{-0.1cm}
\caption{Motivation of the proposed CapHDR2IR model.}
\vspace{-0.5cm}
\end{figure}

Infrared (IR) imaging is a powerful tool with a wide range of applications, offering unique advantages in various fields such as surveillance \cite{surveillance}, medical diagnostics \cite{medical_diagnostics}, and intelligent transportation systems \cite{Infrared3}.
This is primarily because IR images can capture clear images in poor visibility conditions, such as at night \cite{DBR, Infrared1}.
This capability ensures that critical details are captured even in low-light or adverse weather situations, enhancing situational awareness and safety in many scenarios \cite{Infrared2, CourtNet}.

Obtaining high-quality infrared images poses significant challenges.
The equipment required for capturing IR images is often expensive and has limitations in terms of resolution and sensitivity \cite{low_resolution}.
High-resolution IR cameras can be prohibitively costly, and lower-cost options typically suffer from poor image quality, making it difficult to utilize IR imaging effectively in many practical applications.
Potential solutions \cite{sRGB-TIR,RGB2IR} have leveraged standard dynamic range (SDR) data from the visible light spectrum to generate infrared images.
This method allows for the creation of IR images without the need for expensive IR-specific hardware, thereby making IR imaging more accessible and practical.
However, transferring visible light images to the infrared domain poses a significant challenge due to the limited dynamic range of the captured visible content and the absence of scene context. This often results in a loss of fidelity in the reconstructed IR images.
These challenges can be addressed by: 1) providing HDR content from visible light and 2) incorporating scene context through dense captions to provide cues for IR reconstruction.

\textbf{Loss of fidelity due to dynamic range of visible content} arises from the fundamental nature of infrared imaging.
Infrared imaging is typically used to provide supplementary information when visible light is unreliable or unavailable, such as in stark, challenging or low-light conditions \cite{DFNet, SiamIVFN}.
Relying on visible light to generate reliable infrared images can be considered a paradox, as it undermines the inherent advantage of infrared imaging.
This dependency on visible light to produce reliable infrared imagery is problematic, limiting its practical applications in scenarios where visible light is inherently insufficient.
An example can be seen in Fig. \ref{fig:Comparison-between-SDR}, where the generated infrared image from the captured SDR image in the dark is not reliable.

High dynamic range (HDR) images, on the other hand, offer a robust and widely adopted solution to the limitations of standard dynamic range (SDR) images.
HDR images capture a broader range of luminosity levels, enabling the visualization of details in both the darkest and brightest parts of a scene.
As shown in Fig. \ref{fig:Comparison-between-SDR}, dark building edges cannot be seen clearly in the SDR image.
However, the HDR image contains more information, which has multiple exposure blankets.
In this example, the building edges are clear at EV +3, and the objects inside the buildings are clear at EV -3.
The robustness of HDR enhances its reliability, making it a better choice for transferring IR images.

\textbf{Loss of fidelity caused by insufficient contextual awareness} manifests as pseudo-thermal-crossover artifacts in generated infrared images.
Due to the different wavelengths, the content presented in visible light images can significantly differ from that in infrared images.
In particular, thermal crossover occurs when objects with similar temperatures appear indistinguishable in infrared images.
For instance, in the forest scenery from Fig. \ref{fig:Vision-language-model-is}, visible light images can clearly differentiate tree trunks and branches, but these elements may appear very similar in infrared due to their similar temperatures.
With a huge amount of these kinds of training data, generator models are prone to converting visible light images into ``blurry gray blobs'' that simulate thermal crossover, making the images less informative.
For example, in Fig. \ref{fig:Vision-language-model-is}, the generated infrared image (w/o caption) cannot distinguish useful details, such as the people from the background, while the real infrared image can easily achieve this.
Consequently, this issue limits the utility of converting visible light to infrared images for practical applications.

Currently, vision-language models (VLMs) exhibit remarkable capabilities in content recognition and understanding \cite{vision_language_model1}.
They are adept at identifying and interpreting intricate scenes, offering a thorough and detailed understanding of image content.
By integrating visual and textual information, VLMs can achieve superior performance in tasks such as image captioning, navigation, and scene understanding \cite{vision_language_model2}.
Compared to direct pixel-level domain transfer, a potential approach involves using VLMs to understand the content of the images before generating infrared images.
By understanding the context and content of the visible light images, the generator can generate infrared images that retain meaningful information, avoiding producing meaningless blurry gray regions.

Motivated by the above analysis, we propose a novel transferring visible-to-infrared method called CapHDR2IR in this paper.
To handle the reliability problem, CapHDR2IR utilizes HDR images to generate infrared images, ensuring that the converted images maintain high fidelity and reliability even in challenging conditions.
To avoid generating pseudo-thermal-crossover artifacts, CapHDR2IR combines a dense captioning branch to provide a semantic and fine-grained understanding of the image content. 
This entails that the generated infrared images are meaningful and retain critical information.

In summary, our contributions are summarized below:
\begin{enumerate}
\item To the best of our knowledge, we are the first to introduce HDR images to enhance the reliability of generating infrared images from visible light images.
\item To ensure that the generated infrared images are informative and meaningful, we are the first to integrate VLMs into the image generation task.
\item Our approach achieves state-of-the-art performance on the HDRT dataset, demonstrating significant improvements in the quality and reliability of the proposed CapHDR2IR model.
\end{enumerate}

%% file: Sec2_Related_Works.tex
\section{Related Works}

\subsection{Domain Transfer and RGB-to-IR}

Domain transfer, also known as domain adaptation, can transfer data from one domain to another.
It is employed in various applications, including style transfer, image synthesis, and modality conversion.
Traditional methods like histogram matching \cite{HM1, HM2}, color transfer \cite{ColorTransfer}, and K-nearest neighbor \cite{KNN} adjust the intensity distribution of an input image to match the statistical properties of a target image.
While effective for linear domain transfer tasks, these methods fall short in non-linear transfers, such as converting visible light to infrared.

\begin{figure*}[!t]
\vspace{-0.3cm}
\centering
\includegraphics[width=0.8\textwidth]{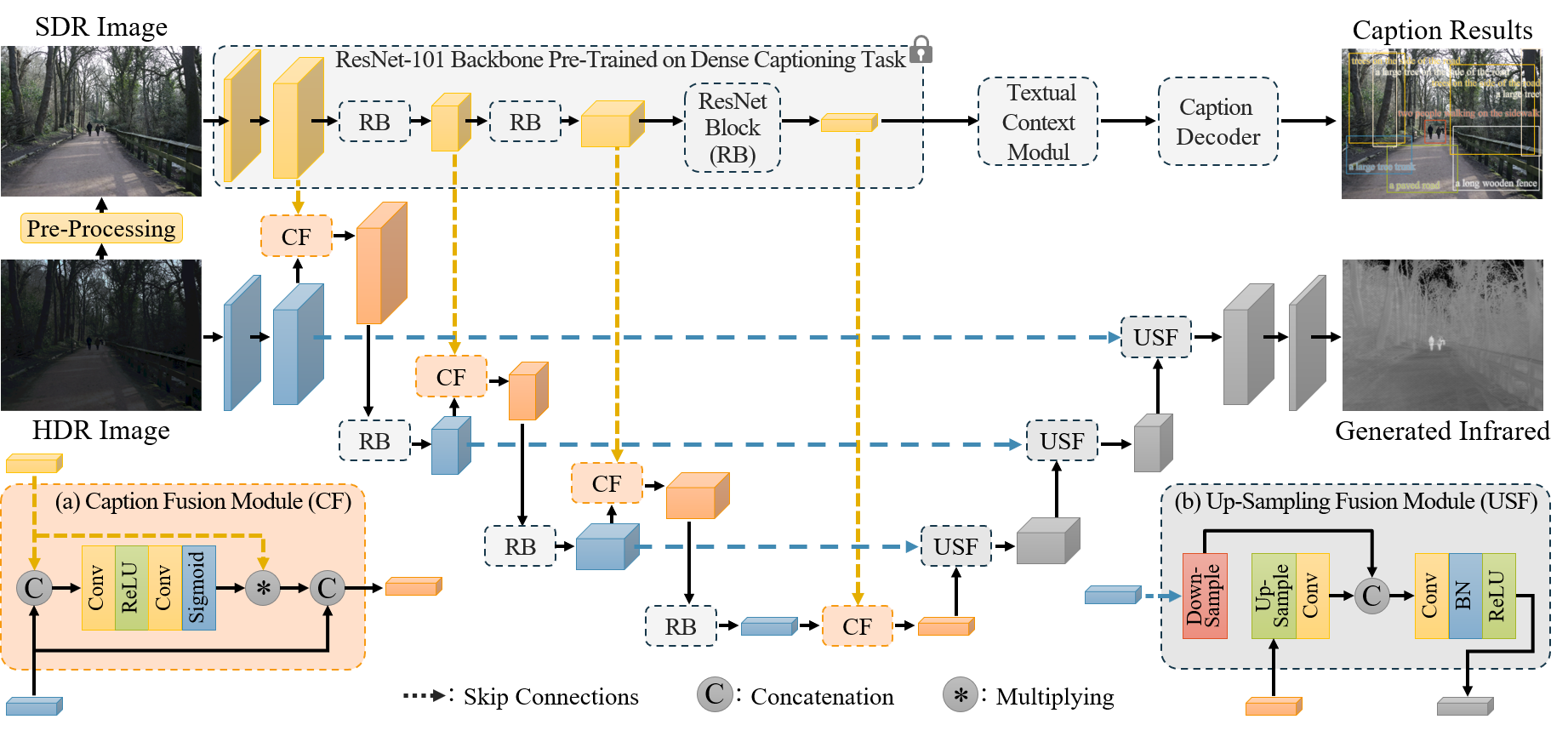}
\vspace{-0.2cm}
\caption{\label{fig:The-overall-structure}The overall structure of the proposed CapHDR2IR.
The process comprises two branches: a caption branch (upper part) and an image generation branch (lower part).
The caption branch adopts a ResNet-101 backbone pre-trained on dense captioning to comprehend the image context. 
In the IR image generation branch, the encoder stage integrates features with contextual information through caption fusion modules (CF), while the decoder stage refines and upscales these features by up-sampling fusion modules (USF) to generate infrared images that retain both high dynamic range content and scene context.
}
\vspace{-0.3cm}
\end{figure*}

Many deep-learning-based methods have been developed for visible light to infrared image translation.
PAS-GAN \cite{PAS-GAN} employed a lightweight pyramid across-scale feature extraction module and a gradient loss function.
InfraGAN \cite{InfraGAN} introduced SSIM and pixel-based loss in its discriminator.
IR-GAN \cite{IR-GAN} featured ConvNeXt architecture and gradient vector loss.
The sRGB-TIR \cite{sRGB-TIR} employed edge-guided multi-domain translation to improve the accuracy of generating thermal infrared images from RGB inputs. 

These methods have found numerous applications across various fields.
For example, Huang et al. \cite{RGB2IR} enhanced the computation of optical flow in infrared images by leveraging cross-modal image generation from RGB inputs.
IC-GAN \cite{IC-GAN} has been effectively applied to forest fire monitoring.
In agricultural applications, infrared image predictions from RGB images achieved state-of-the-art performance in plant phenotyping \cite{HR-NIR, RGB-NIR}.

However, existing methods still do not perform well under low light conditions or other challenging
environments, and they are prone to generating pseudo-thermal-crossover artifacts.
High dynamic range (HDR) imaging addresses this problem due to its capability to capture a broader range of luminance levels. In photography and videography, HDR techniques have significantly enhanced image detail and visual quality in both bright and dark areas \cite{HDR_bright, HDR_dark}.
In the medical field, HDR has improved contrast and visibility in diagnostic images, aiding better diagnosis \cite{HDR_medical}.
Automotive industries have employed HDR cameras in advanced driver-assistance systems (ADAS) for improved visibility under challenging lighting conditions \cite{HDR_ADAS}.
Additionally, HDR is critical in computer vision for tasks such as object recognition and scene reconstruction \cite{HDR_datection}, demonstrating its extensive applicability and impact.

\subsection{Vision-Language Models and Dense Captioning}

Vision-language models (VLMs) represent a novel deep-learning paradigm inspired by advances in natural language processing.
VLMs operate on a vision-language model pre-training and zero-shot prediction paradigm.
In this approach, a VLM is pre-trained with large-scale image-text pairs, which are abundantly available on the internet.
These pre-trained models can then be directly applied to downstream visual recognition tasks without the need for further fine-tuning.
The pre-training process is guided by specific vision-language objectives, enabling the model to learn image-text correspondences from the vast dataset of image-text pairs.
Among these applications, dense captioning provides detailed descriptions for significant image regions in natural language, facilitating various tasks such as blind navigation, human-robot interaction, and autonomous driving.
Dense captioning benefits greatly from the capabilities of VLMs, as they can understand and describe complex scenes effectively, enhancing the performance and applicability of dense captioning methods in real-world scenarios.

Johnson et al. combined a region proposal network (RPN) with an LSTM to identify and generate captions for regions of interest (ROIs) \cite{caption_relatedwork1}.
However, they ignored additional context, treating each RoI independently.
Many follow-ups improved dense captioning performance by integrating contextual information. Yang et al. combined region features with global context features for coarse context \cite{caption_relatedwork2}.
Other methods aimed to capture finer contextual details. Yin et al. used a non-local similarity graph for caption generation \cite{caption_relatedwork3};
Li et al. employed object detection to assist the model training \cite{caption_relatedwork4};
and Shao et al. introduced a specialized loss module to enhance region-object
correlation \cite{caption_relatedwork5}. 

Among them, ETDC \cite{CapModel} captured surrounding textual context and balanced word frequencies by a re-sampling strategy during training, enhancing language learning efficiency.
ETDC has demonstrated superior performance on dense captioning datasets, surpassing the state-of-the-art in mean average precision.
By adopting ETDC as the captioning branch, our model ensures that the generated infrared images are enriched with detailed and meaningful descriptions, providing a robust foundation for high-quality image generation.

%% file: Sec3_Methodology.tex
\section{Methodology}

Our approach addresses the reliability in infrared images derived from HDR images by integrating a VLM branch pre-trained on dense captioning with an image generator.
This section first overviews the proposed CapHDR2IR and then introduces the main modules.

\subsection{Overall Structure}

The architecture of our model consists of two branches: a caption branch and an image generation branch, as seen in Fig. \ref{fig:The-overall-structure}.
The caption branch aims to distill the content of the image through dense caption features.
The image generation branch fuses the caption information extracted from the caption branch and focus on infrared image generation.

The image generation branch comprises an encoder and a decoder.
The encoder processes the input images, producing visual features.
With the caption fusion modules, the encoder incorporates information from the caption branch, which includes contextual features about the scene.
Both the encoder and the caption fusion module operate at multiple scales to capture a comprehensive range of features from coarse to fine.

The decoder consists of several up-sampling modules that increase the resolution of the feature maps while fusing multi-scale features.
At each stage, features from the previous layer and the same-scale encoder are fused by up-sampling fusion modules.
The final stage adjusts the feature maps to the target resolution and applies a sigmoid activation to map the output values from 0 to 1, generating the final infrared images.
The details of the caption branch, caption fusion module, and up-sampling fusion module are introduced in the following sub-sections.

During the training phase, we employ a combination of perceptual loss $\mathcal{L}_{\textrm{per}}$ and GAN loss $\mathcal{L}_{GAN} $:
\begin{equation}
\mathcal{L_{\textrm{total}}}=\alpha\mathcal{L}_{\textrm{per}}+\beta\mathcal{L}_{GAN},
\end{equation}
where the weights $\alpha=10$ and $\beta=0.1$ were decided through experiments.
Perceptual loss ensures that the generated infrared images capture high-level features and semantic content similar to the ground truth images.
By comparing activations from a pre-trained VGG network, perceptual loss helps produce visually coherent results that maintain the overall structure and details of the scene.
GAN loss is to make the generated infrared images indistinguishable from real infrared images, which makes them realistic and high-quality.
By combining these loss functions, our model achieves perceptually accurate and realistic infrared image generation.

\subsection{Caption Branch}

Since dense captioning models commonly use SDR images, the caption branch first converts HDR images into SDR images by the log-average function.
The core idea of this method is to adjust the overall brightness of the image using the logarithmic average luminance.
Given an HDR image, where each pixel has a luminance value $L_{HDR}(i,j)$, the log-average luminance $\bar{L}_{HDR}$ is calculated as: 
\begin{equation}
\bar{L}_{HDR}=\exp\left(\frac{1}{N}\sum_{i,j}\log\left(\epsilon+L_{HDR}(i,j)\right)\right),
\end{equation}
where $N$ is the total number of pixels in the image, and $\epsilon$ is a small constant to avoid issues with the logarithm of zero.
The luminance value of SDR images $L_{SDR}(i,j)$ can be obtained by using the log-average luminance for each pixel luminance: 
\begin{equation}
\bar{L}_{SDR}(i,j)=\frac{\alpha\cdot L_{HDR}(i,j)}{\bar{L}_{HDR}},
\end{equation}
where $\alpha$ is a scaling factor to control the brightness level of the final image.
To prevent excessively high luminance values, further compression is implemented: 
\begin{equation}
L_{SDR}(i,j)=\frac{\bar{L}_{SDR}(i,j)}{1+\bar{L}_{SDR}(i,j)}.
\end{equation}

After obtaining the SDR images, inspired by \cite{CapModel}, the caption branch employs a ResNet-101 backbone and an RPN to extract RoI features with their coordinates.
To enhance contextual information, the textual context module (TCM) \cite{CapModel} layers are preserved, and the visual features are fed into the caption decoder to generate descriptive sentences for each RoI.
For more details, please refer to \cite{CapModel}.

\subsection{Caption Fusion Module}

The caption fusion module enhances understanding of scene content by integrating features from the caption branch with the same scale visual features from the encoder.
This module ensures that the generated infrared images are both visually and semantically meaningful.
To integrate caption features with visual features effectively, we use a spatial attention module, which generates attention maps that highlight relevant regions in the caption features, enhancing the fusion process.
Fig.~\ref{fig:The-overall-structure}a elaborates the structure of the caption fusion module.
The first step is to calculate the attention map:
\begin{equation}
Att\_map=\sigma(\text{Conv}(\text{ReLU}(\text{Conv}(\text{Cat}(F_{\textrm{vis}}^{i},F_{\textrm{cap}}^{i})))))
\end{equation}
where $F_{\textrm{vis}}^{i}$ and $F_{\textrm{cap}}^{i}$ are input features from the encoder of the image generation branches and caption branches.
$\text{Cat}(\cdot)$ denotes concatenation.
$\text{Conv}(\cdot)$ is a convolutional layer.
$\text{ReLU}(\cdot)$ is a rectified linear unit function, and $\sigma(\cdot)$ is the sigmoid function.
The caption features are then element-wise multiplied by the attention map, aligning them with the visual features to ensure consistency:
\begin{equation}
F_{\textrm{aligned\_cap}}^{i}=Att\_map\cdot F_{\textrm{cap}}^{i}.
\end{equation}
Finally, the aligned caption features $F_{\textrm{aligned\_cap}}^{i}$ and the visual features $F_{\textrm{vis}}^{i}$ are concatenated to get the the current output:
\begin{equation}
\text{\ensuremath{F_{vis}^{i+1}}=Cat}(F_{\textrm{fused\_cap}}^{i},F_{\textrm{vis}}^{i}).
\end{equation}
The caption fusion process is progressively integrated at multiple scales, capturing comprehensive scene information.

\begin{table*}[!t]
\caption{\label{tab:The-results-of}The results of the HDRT dataset.
The best results are highlighted with \textcolor{red}{red}, and the second-best results with \textcolor{orange}{orange}.
The table is divided into two subgroups: 1) taking SDR images as the input and 2) taking HDR images as the input.
``$\uparrow$'' indicates that a higher value for this metric corresponds to better performance, and vice versa for ``$\downarrow$''.
}
\vspace{-0.3cm}
\centering{}
\resizebox{\textwidth}{!}{
\begin{tabular}{clcccc}
\toprule 
Input &  Methods  &  PSNR $\uparrow$ (\texttimes 10)  &  SSIM $\uparrow$ (\texttimes 0.1)  &  MSE $\downarrow$ (\texttimes 0.1)  &  LPIPS $\downarrow$ (\texttimes 0.1) 
\tabularnewline
\midrule
\multirow{8}{*}{ SDR }
 &  Histogram Matching \cite{HM2,HM1}  &  0.973  &  2.073  &  6.221  &  4.851 
\tabularnewline
 &  Color Transfer \cite{ColorTransfer}  &  1.134  &  4.150  &  5.814  &  5.632 
\tabularnewline
 &  CycleGAN \cite{CycleGAN}  &  1.082  &  5.098  &  4.149  &  4.689 
\tabularnewline
 &  Pix2Pix \cite{Pix2Pix}  &  1.072  &  5.041  &  4.041  &  4.649 
\tabularnewline
 &  MUNIT \cite{MUNIT}  &  1.409  &  4.578  &  3.886  &  5.823 
\tabularnewline
 &  sRGB-TIR \cite{sRGB-TIR}  &  1.202  &  3.691  &  7.858  &  4.624 
\tabularnewline
 &  RGB2IR \cite{RGB2IR}  &  \textcolor{orange}{1.645}  &  \textcolor{orange}{5.152}  &  \textcolor{orange}{3.291}  &  \textcolor{orange}{4.327} 
\tabularnewline
 &  \textbf{CapHDR2IR (Ours)}  &  \textbf{\textcolor{red}{1.879}}  &  \textbf{\textcolor{red}{6.163}}  &  \textbf{\textcolor{red}{2.574}}  &  \textbf{\textcolor{red}{3.141}} 
\tabularnewline
\midrule
\multirow{8}{*}{ HDR }
 &  Histogram Matching \cite{HM2,HM1}  &  1.023  &  2.174  &  6.138  &  4.613 
\tabularnewline
 &  Color Transfer \cite{ColorTransfer}  &  1.243  &  4.147  &  5.003  &  5.419 
\tabularnewline
 &  CycleGAN \cite{CycleGAN}  &  1.057  &  5.151  &  4.094  &  4.547 
\tabularnewline
 &  Pix2Pix \cite{Pix2Pix}  &  1.066  &  5.138  &  4.060  &  4.730 
\tabularnewline
 &  MUNIT \cite{MUNIT}  &  1.386  &  4.889  &  3.913  &  5.914 
\tabularnewline
 &  sRGB-TIR \cite{sRGB-TIR}  &  1.495  &  5.019  &  3.332  &  4.528 
\tabularnewline
 &  RGB2IR \cite{RGB2IR}  &  \textcolor{orange}{1.750}  &  \textcolor{orange}{5.530}  &  \textcolor{orange}{2.756}  &  \textcolor{orange}{4.066} 
\tabularnewline
 &  \textbf{CapHDR2IR (Ours)}  &  \textbf{\textcolor{red}{1.976}}  &  \textbf{\textcolor{red}{6.359}}  &  \textbf{\textcolor{red}{2.242}}  &  \textbf{\textcolor{red}{3.035}} \tabularnewline\bottomrule 
\end{tabular}} 
\vspace{-0.3cm}
\end{table*}

\subsection{Up-Sampling Fusion Module}

The up-sampling fusion module, shown in Fig.~\ref{fig:The-overall-structure}b, progressively reconstructs infrared images from multi-scale feature maps.
It integrates features from lower-resolution stages and higher-resolution details to produce a coherent and detailed output.
The module includes an up-sampling block, a down-sampling block, and a convolution block.
The up-sampling block uses transpose convolution to double the spatial dimensions of the features from the previous layer, which are then fed into a convolution layer:
\begin{equation}
F_{\textrm{up}}^{i}=\text{Conv}(\text{ConvTrans}(F_{\textrm{de}}^{i}),
\end{equation}
where $\text{ConvTrans}(\cdot)$ stands for the transposed convolution operator.
To ensure that the features from the encoder layer and the up-sampled decoder features align correctly, a center cropping operation is applied to the encoder features  $F_{\textrm{vis}}^{i}$.
This operation adjusts the dimensions of the skip connection features to match those of the up-sampled features.

The cropped encoder features $F_{\textrm{vis}}^{i}$ are concatenated with the up-sampled features $F_{\textrm{up}}^{i}$, combining low-level spatial details with high-level semantic information.
The fused features are processed through a series of convolutional layers followed by batch normalization ($\text{BN}(\cdot)$) and ReLU activation to refine the fused features and produce the final output for the current stage:
\begin{equation}
F_{\textrm{de}}^{i+1}=\text{ReLU}(\text{BN}(\text{Conv}(\text{Cat}(F_{\textrm{vis}}^{i},F_{\textrm{up}}^{i}))))
\end{equation}

This process is repeated for each stage of the decoder, progressively reconstructing the infrared image from the multi-scale features.

%% file: Sec4_Experiments.tex
\section{Experiments}

In this section, we first introduce experimental settings, the datasets, and evaluation metrics.
We subsequently compare our CapHDR2IR model with other methods.
Finally, ablation studies are presented to highlight the performance improvement of our main modules.

\subsection{Implementation Details}

Our experiments were conducted using PyTorch equipped with an Intel Xeon@2.5GHz CPU and an NVIDIA L40 GPU.
For training, we used a batch size of 16 and trained the model for 200 epochs.
The learning rate was set to $4\times10^{-5}$ for both the generator and discriminator.
We employed the Adam optimizer for both networks, with $\beta_{1}=0.9$ and $\beta_{2}=0.99$.
The loss weights were set as follows: 10.0 for perceptual loss and 0.1 for GAN loss.\\\vspace{-0.3cm}

\noindent \textbf{Dataset}.
We evaluated our model on the HDRT dataset \cite{HDRTDataset}, which includes 10,000 image pairs.
The dataset covers various scenes such as urban, rural, day, and night environments.
Each pair of images provides a high dynamic range visible light image and its well-registered infrared image.
The training set contains 8,000 image pairs, and the test set contains 2,000 image pairs.\\\vspace{-0.3cm}

\noindent \textbf{Evaluation Metrics}.
To evaluate the performance of our model, we used several evaluation metrics:
\begin{itemize}
\item Peak Signal-to-Noise Ratio (PSNR): Measures the ratio between the maximum possible power of a signal and the power of corrupting noise that affects the fidelity of its representation.
\item Structural Similarity Index (SSIM): Evaluates the perceived quality of the generated images by comparing structural information.
\item Mean Squared Error (MSE): Calculates the average squared difference between the generated images and the ground truth.
\item Learned Perceptual Image Patch Similarity (LPIPS): Uses a deep network to measure perceptual similarity, focusing on human perceptual judgments.
\end{itemize}

\subsection{Comparison with Other Methods}

\begin{figure*}[!t]
\vspace{-0.3cm}
\includegraphics[width=1\textwidth]{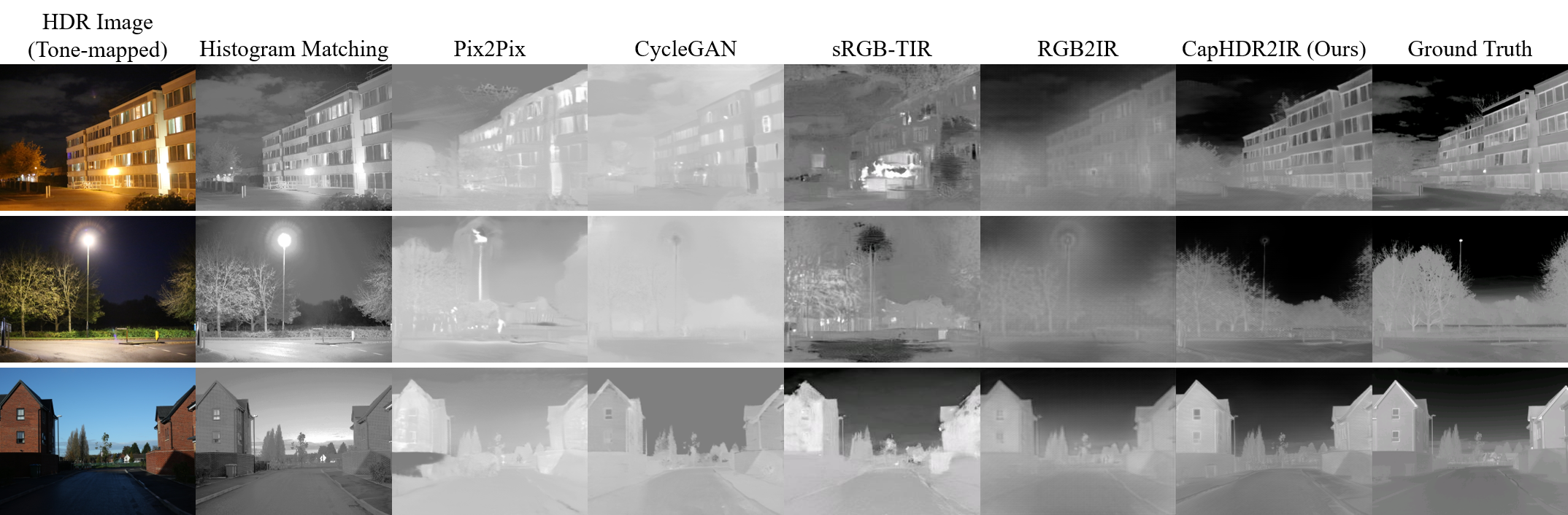}
\par
\vspace{-0.2cm}
\caption{\label{fig:Visualization-results-on}Visualization results on the HDRT dataset.}
\vspace{-0.2cm}
\end{figure*}

\begin{table*}[!t]
\caption{\label{tab:Ablation-study-of}Ablation study on the HDRT dataset. 
``\textsurd'' indicates the technique is adopted, ``\texttimes'' indicates not, and ``-'' means not applicable.
``$\uparrow$'' indicates that a higher value for this metric corresponds to better performance, and vice versa for ``$\downarrow$''.
}
\vspace{-0.3cm}
\centering{}
\resizebox{\textwidth}{!}{
\begin{tabular}{lcccccccc}
\toprule
Methods 
 &  
\begin{tabular}{@{}c@{}}
HDR\tabularnewline
Input\tabularnewline
\end{tabular} 
 &  
\begin{tabular}{@{}c@{}}
Pre-\tabularnewline
Processing\tabularnewline
\end{tabular} 
 &  
\begin{tabular}{@{}c@{}}
Caption\tabularnewline
Branch\tabularnewline
\end{tabular} 
 &  
\begin{tabular}{@{}c@{}}
Caption\tabularnewline
Fusion\tabularnewline
\end{tabular} 
 &  
PSNR $\uparrow$ (\texttimes 10) 
 &  
SSIM $\uparrow$ (\texttimes 0.1) 
 &  
MSE $\downarrow$ (\texttimes 0.1) 
 &  
LPIPS $\downarrow$ (\texttimes 0.1) 
\tabularnewline
\midrule
 SDR2IR\_Baseline  &  \texttimes{}  &  -  &  \texttimes{}  &  -  &  1.796  &  5.904  &  2.982  &  3.362 
\tabularnewline
 SDR2IR\_V1  &  \texttimes{}  &  -  &  \textsurd{}  &  \texttimes{}  &  1.864  &  6.126  &  2.723  &  3.149 
\tabularnewline
 SDR2IR\_V2  &  \texttimes{}  &  -  &  \textsurd{}  &  \textsurd{}  &  1.879  &  6.162  &  2.574  &  3.141 
\tabularnewline
\midrule
 HDR2IR\_Baseline  &  \textsurd{}  &  \texttimes{}  &  \texttimes{}  &  -  &  1.913  &  6.177  &  2.386  &  3.215 
\tabularnewline
 HDR2IR\_V1  &  \textsurd{}  &  \texttimes{}  &  \textsurd{}  &  \texttimes{}  &  1.945  &  6.303  &  2.294  &  3.106 
\tabularnewline
 HDR2IR\_V2  &  \textsurd{}  &  \texttimes{}  &  \textsurd{}  &  \textsurd{}  &  1.955  &  6.342  &  2.281  &  3.088 
\tabularnewline
 HDR2IR\_V3  &  \textsurd{}  &  \textsurd{}  &  \textsurd{}  &  \texttimes{}  &  \textcolor{orange}{1.966}  &  \textcolor{red}{6.359}  &  \textcolor{orange}{2.260}  &  \textcolor{orange}{3.057} 
\tabularnewline
 \textbf{CapHDR2IR (Ours)}  &  \textbf{\textsurd{}}  &  \textbf{\textsurd{}}  &  \textbf{\textsurd{}}  &  \textbf{\textsurd{}}  &  \textbf{\textcolor{red}{1.976}}  &  \textbf{\textcolor{red}{6.359}}  &  \textbf{\textcolor{red}{2.242}}  &  \textbf{\textcolor{red}{3.035}} 
\tabularnewline
\bottomrule
\end{tabular}} 
\vspace{-0.3cm}
\end{table*}

We compared CapHDR2IR with existing methods, which include both traditional image transfer approaches and state-of-the-art deep learning-based techniques.
The methods compared include: traditional histogram matching \cite{HM2,HM1} and color transfer \cite{ColorTransfer}, which are designed to enhance visual similarity between the images;
CycleGAN \cite{CycleGAN}, a popular image-to-image translation framework based on Generative Adversarial Networks (GANs) that learn mappings between two domains without paired examples;
Pix2Pix \cite{Pix2Pix}, an image-to-image translation model that uses paired datasets and is based on conditional GANs;
and its improvement designed for transferring visible light to infrared, RGB2IR \cite{RGB2IR};
MUNIT \cite{MUNIT}, which transfers images from one domain to another by learning disentangled representations of content and style;
and its improvement designed for transferring visible light to infrared, sRGB-TIR \cite{sRGB-TIR}.

Table \ref{tab:The-results-of} summarizes the performance of our CapHDR2IR model compared to other methods on the HDRT dataset.
The table is divided into two parts: taking SDR images as input and taking HDR images as input, with the same methods used for generating infrared images for comparison.
Results show that HDR inputs consistently outperform SDR inputs across all methods, highlighting the advantage of HDR images for generating infrared images.
Additionally, our model surpasses traditional methods and other deep learning approaches by effectively leveraging high-level semantics and detailed visual features through dense captioning integration.
This enhances performance, particularly in maintaining the semantic integrity and visual coherence of the generated infrared images.
Consequently, our method achieves the best performance across all metrics in both SDR and HDR groups.

Fig. \ref{fig:Visualization-results-on} gives qualitative results, showcasing various challenging scenarios, including over-exposed conditions, low-light conditions, and complex backgrounds.
As observed, our approach successfully preserves important scene details and object boundaries, which are often lost in other RGB-to-IR conversion methods.
The integration of HDR imaging further enhances the robustness of our model, allowing it to handle a wider range of luminosity levels and maintain visual fidelity.
Specifically, the generated infrared images exhibit superior texture details and sharper edges compared to other methods.
In low-light conditions, our model effectively highlights critical features such as buildings and roads, which are crucial for applications like surveillance and autonomous driving.
Additionally, our method demonstrates excellent performance in distinguishable objects by leveraging the caption branch to understand the visible light image.
This qualitative comparison highlights the ability of our model to produce infrared images with reduced noise and artifacts, contributing to clearer and more reliable visual outputs.

\begin{figure*}[!t]
\vspace{-0.2cm}
\includegraphics[width=1.0\textwidth]{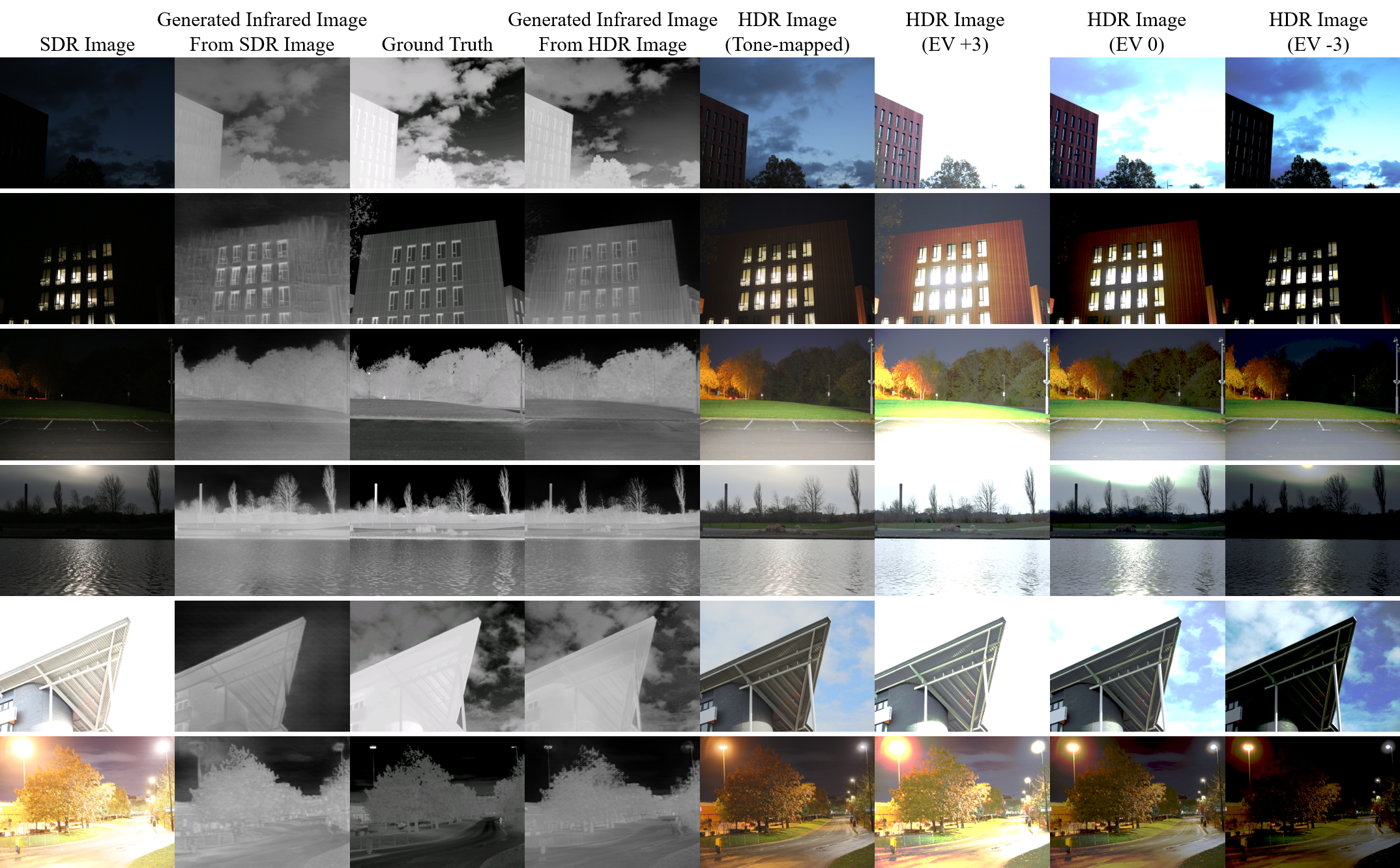}
\par
\vspace{-0.2cm}
\caption{\label{fig:HDR_vs_SDR}Comparison of generation performance using SDR vs. HDR inputs.
}
\end{figure*}

\subsection{Ablation Study}

In this section, we conduct a comprehensive ablation study to evaluate the impact of different components of our proposed CapHDR2IR model.
The ablation study examines the following aspects: 1) the use of HDR input, 2) the inclusion of pre-processing in the caption branch, 3) the presence of the caption branch, and 4) the use of the caption fusion module.
Table \ref{tab:Ablation-study-of} presents the quantitative results of various configurations, highlighting the contributions of each component to the overall performance.\\\vspace{-0.3cm}

\noindent \textbf{HDR Input}.
To investigate the effect of HDR input on the model's performance, we compare the results of using HDR images versus SDR images.
The configurations labeled as ``SDR2IR\_Baseline'', ``SDR2IR\_V1'', and ``SDR2IR\_V2'' utilize SDR inputs, whereas ``HDR2IR\_Baseline'', ``HDR2IR\_V1'', ``HDR2IR\_V2'', ``HDR2IR\_V3'', and ``CapHDR2IR'' employ HDR inputs.
As observed, the models with HDR input consistently outperform their SDR counterparts in all metrics, particularly in terms of PSNR and SSIM, demonstrating the advantage of leveraging HDR imaging for generating high-quality infrared images.
\\\vspace{-0.3cm}

\noindent \textbf{Pre-Processing in Caption Branch}.
HDR2IR\_Baseline excludes pre-processing, while HDR2IR\_V1, HDR2IR\_V2, and CapHDR2IR include it. Results show that pre-processing helps the caption branch capture richer semantic features from HDR images.\\\vspace{-0.3cm}

\noindent \textbf{Caption Branch}.
SDR2IR\_Baseline and HDR2IR\_Baseline do not incorporate the caption branch, whereas SDR2IR\_V1, SDR2IR\_V2, HDR2IR\_V1, HDR2IR\_V2, HDR2IR\_V3, and CapHDR2IR include it.
Results show that the presence of the caption branch significantly boosts performance across all metrics.
This highlights the role of the caption branch in providing essential contextual information that enhances the semantic understanding of the scene.
\\\vspace{-0.3cm}

\noindent \textbf{Caption Fusion Module}.
SDR2IR\_V1, HDR2IR\_V1, and HDR2IR\_V2 exclude the caption fusion module, while SDR2IR\_V2, HDR2IR\_V3, and CapHDR2IR incorporate it.
The inclusion of the caption fusion module leads to further improvements, particularly in SSIM and LPIPS, indicating that it plays a crucial role in merging semantic information from the caption branch with visual features, thus enhancing the overall image quality.\\\vspace{-0.3cm}

\noindent \textbf{Loss Weight}

In order to test the impact of different weights $\alpha$ and $\beta$ on IR image generating performance, we use CapHDR2IR with different weights to train and reconstruct IR images; the results are shown in Tab. \ref{tab:Ablation-alpha-beta}.
The results show that the performance is best when $\alpha$ is set to 10 and $\beta$ is set to 0.1. 

\begin{table}[!h]
\caption{\label{tab:Ablation-alpha-beta} Ablation study on $\alpha$ and $\beta$.}
\centering{}\resizebox{\columnwidth}{!}{
\begin{tabular}{cccccc}
\toprule
$\alpha$ & $\beta$ & PSNR $\uparrow$ (\texttimes 10)  &  SSIM $\uparrow$ (\texttimes 0.1)  &  MSE $\downarrow$ (\texttimes 0.1)  &  LPIPS $\downarrow$ (\texttimes 0.1) \tabularnewline
\midrule 
e-1 & e-1 & 1.526  & 4.939 & 3.525 & 4.479\tabularnewline
e-1 & 1   & 1.449  & 4.776 & 4.202 & 4.783\tabularnewline
e-1 & e+1 & 1.455  & 4.472 & 4.306 & 4.735\tabularnewline
1   & e-1 & 1.504  & 5.179 & 4.332 & 4.578\tabularnewline
1   & 1   & 1.476  & 4.967 & 4.340 & 4.991\tabularnewline
1   & e+1 & 1.460  & 4.823 & 4.674 & 5.060\tabularnewline
e+1 & e-1 &  \textbf{1.879}  &  \textbf{6.163}  &  \textbf{2.574}  &  \textbf{3.141} \tabularnewline
e+1 & 1   & 1.502  & 5.028 & 4.326 & 4.223\tabularnewline
e+1 & e+1 & 1.491  & 4.865 & 3.865 & 4.591\tabularnewline
\bottomrule 
\end{tabular}}
\end{table}

\subsection{Visual Results}

This section provides qualitative results by pruning two key contributions, HDR input and the caption branch, to intuitively demonstrate the effectiveness of our CapHDR2IR.

\noindent \textbf{HDR Input vs. SDR Input} 

To demonstrate the advantage of the HDR images as the input, we compared infrared images generated from both SDR and HDR inputs, as shown in Fig. \ref{fig:HDR_vs_SDR}. 
The comparison shows that SDR-generated infrared images often miss details due to their sensitivity to varying light conditions.

In underexposed cases, for instance, in the first row of Fig. \ref{fig:HDR_vs_SDR}, the SDR image struggles with the building, losing information that is effectively retained in HDR-generated images, which can be extracted from different exposure levels (e.g., EV +3 for building structures and EV -3 for clouds).
Similarly, in the second row, SDR images lose structural details around building edges due to the low-light condition.
In contrast, HDR images maintain a balanced contrast between bright windows and darker facades, demonstrating the advantage of HDR inputs in preserving such details.
In the third row, SDR images show poor distinction between the concrete floor, grass, and distant trees.
In contrast, HDR images provide clear separation and improved visibility, which is particularly noticeable in the tone-mapped HDR outputs. 

Further, the fourth row illustrates that the limited dynamic range of SDR images results in a lack of clarity in depicting plants and sunlight simultaneously.
Therefore, the IR image generated from the SDR image lacks depth and clarity of them.
However, the HDR-generated IR image preserves these critical details.
The fifth row highlights the limitation of SDR inputs in capturing cloud details in an overexposed sky.
HDR inputs effectively address this problem, especially with the sky information from EV+3, which reveals cloud details in the resulting infrared image.
In the last row, SDR images exhibit the problem of overexposure around light bulbs, resulting in a loss of detail in its generated IR image.
But HDR images balance these extremes more effectively, capturing both bright and dark areas with minimal loss of detail around light sources. 

Overall, these results clearly demonstrate that HDR inputs lead to superior infrared image generation compared to SDR inputs.
HDR images, by capturing a broader range of luminance, preserve both highlight and shadow details that are often lost with SDR inputs.
This makes HDR inputs particularly valuable in scenes with challenging lighting conditions, where they can produce more reliable and detailed infrared outputs, demonstrating the advantage of leveraging HDR imaging for robust infrared image generation.\\\vspace{-0.3cm}

\begin{figure*}[!t]
\vspace{-0.3cm}
\includegraphics[width=1.0\textwidth]{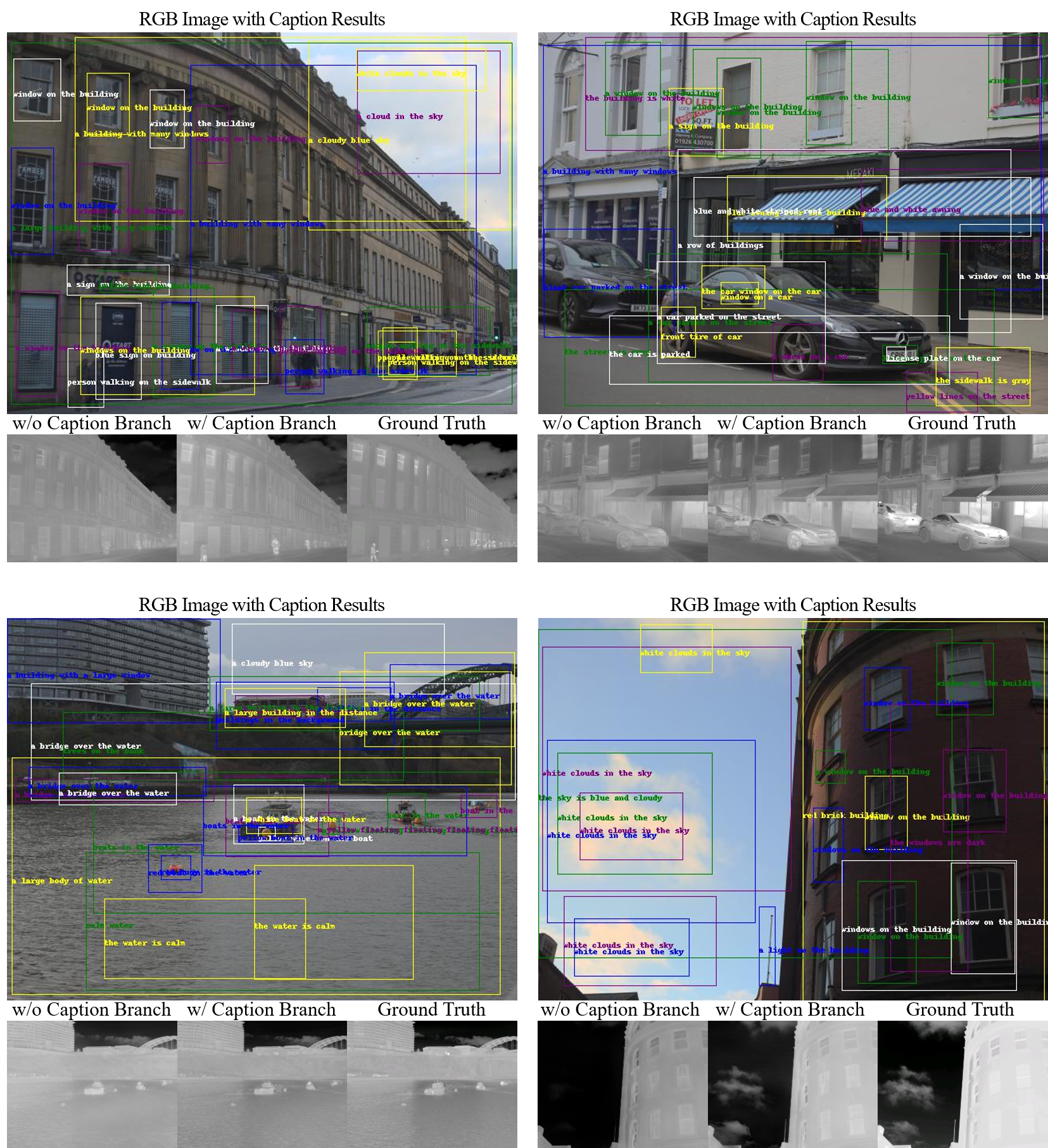}
\vspace{-0.2cm}
\caption{\label{fig:w_wo_cap}Comparison of generation performance with or without the caption branch.
}
\vspace{-0.1cm}
\end{figure*}

\noindent \textbf{w/ Caption Branch vs. w/o Caption Branch} 

To evaluate the impact of incorporating the caption branch on the generating infrared images, we present the visual results shown in Fig. \ref{fig:w_wo_cap}.
In Fig. \ref{fig:w_wo_cap}, the upper row in each pair of images displays the caption results of the corresponding visible light images; the lower row shows the generated infrared images with and without the caption branch and the ground truth.

In the first pair (top left), the visible image is richly annotated with captions describing various scene elements, such as ``window on the building'', ``a cloud in the sky'', and ``people walking on the sidewalk''.
The generated IR image with the caption branch retains these details effectively, preserving the structural integrity of the buildings and the subtle contrasts in the clouds.
In contrast, the infrared image generated without the caption branch fails to capture these details accurately, resulting in losing important scene information, such as people and clouds.

The second pair (top right) shows another example, where the visible image contains detailed captions like ``a car parked on the street''.
With this understanding, the infrared image generated with the caption branch highlights these cars, maintaining the distinct features of the vehicles and the street. 
However, the infrared image generated without the caption branch appears less detailed, with many features blurred or lost, underscoring the importance of semantic understanding of the caption branch.

In the third pair (bottom left), the visible image includes captions such as ``boats in the water'' and ``a large body of water''.
The corresponding infrared image with the caption branch successfully retains the details of the boats and the calm water surface.
Without the caption branch, the infrared image struggles to distinguish between the water and the surrounding structures, leading to pseudo thermal cross artifacts appearing.

In the last pair (bottom right), the visible image captions, including ``white clouds in the sky'', guide the generation of an infrared image that accurately captures clouds.
The absence of the caption branch results in a noticeable absence of clouds.

In summary, Fig. \ref{fig:w_wo_cap} demonstrates that incorporating the caption branch enhances the quality of the generated infrared images.
The caption branch ensures that the model understands and preserves critical scene elements, leading to more detailed and contextually accurate infrared outputs.
This highlights the crucial role of semantic information in bridging the gap between visible and infrared domains, ensuring that the generated images are not only visually like infrared images but also rich in meaningful content.

%% file: Sec5_Conclusion.tex
\section{Conclusion}

In this paper, we presented CapHDR2IR, a novel framework to translate images from the visible light spectrum to the infrared domain,  which effectively addresses the inherent challenges of IR image generation stemming from the dynamic range of input images and contextual awareness.
By leveraging HDR images as inputs, our method solves the dynamic range issue, ensuring more robust and accurate infrared image generation even in challenging lighting conditions, and the incorporation of a dense caption branch allows our model to understand and integrate semantic context from the scene, resulting in more meaningful and contextually accurate infrared outputs.

Future work will incorporate large-scale models to enhance generation capabilities, particularly improving performance in extremely low-light conditions and better generalization across diverse environments.
Additionally, integrating real-time processing and extending the framework to other domain transfer tasks could provide valuable research directions.

%% file: arXiv.bbl
\begin{thebibliography}{39}
\providecommand{\natexlab}[1]{#1}

\bibitem[{Aslahishahri et~al.(2021)Aslahishahri, Stanley, Duddu, Shirtliffe, Vail, Bett, Pozniak, and Stavness}]{RGB-NIR}
Aslahishahri, M.; Stanley, K.~G.; Duddu, H.; Shirtliffe, S.; Vail, S.; Bett, K.; Pozniak, C.; and Stavness, I. 2021.
\newblock From RGB to NIR: Predicting of near infrared reflectance from visible spectrum aerial images of crops.
\newblock In \emph{2021 IEEE/CVF International Conference on Computer Vision Workshops (ICCVW)}, 1312--1322.

\bibitem[{Boroujeni and Razi(2024)}]{IC-GAN}
Boroujeni, S. P.~H.; and Razi, A. 2024.
\newblock IC-GAN: An Improved Conditional Generative Adversarial Network for RGB-to-IR image translation with applications to forest fire monitoring.
\newblock \emph{Expert Systems with Applications}, 238: 121962.

\bibitem[{Castleman(1996)}]{HM2}
Castleman, K.~R. 1996.
\newblock \emph{Digital image processing}.
\newblock USA: Prentice Hall Press.
\newblock ISBN 0132114674.

\bibitem[{de~Lima et~al.(2019)de~Lima, Saqui, Ataky, Jorge, Ferreira, and Saito}]{KNN}
de~Lima, D.~C.; Saqui, D.; Ataky, S.; Jorge, L. A. d.~C.; Ferreira, E.~J.; and Saito, J.~H. 2019.
\newblock Estimating Agriculture NIR Images from Aerial RGB Data.
\newblock In Rodrigues, J. M.~F.; Cardoso, P. J.~S.; Monteiro, J.; Lam, R.; Krzhizhanovskaya, V.~V.; Lees, M.~H.; Dongarra, J.~J.; and Sloot, P.~M., eds., \emph{Computational Science -- ICCS 2019}, 562--574. Cham: Springer International Publishing.
\newblock ISBN 978-3-030-22734-0.

\bibitem[{Gonzalez and Woods(2008)}]{HM1}
Gonzalez, R.; and Woods, R. 2008.
\newblock \emph{Digital Image Processing}.
\newblock Prentice Hall.
\newblock ISBN 9780131687288.

\bibitem[{Haefner et~al.(2021)Haefner, Green, Oursland, Andersen, Goesele, Cremers, Newcombe, and Whelan}]{HDR_dark}
Haefner, B.; Green, S.; Oursland, A.; Andersen, D.; Goesele, M.; Cremers, D.; Newcombe, R.; and Whelan, T. 2021.
\newblock Recovering Real-World Reflectance Properties and Shading From HDR Imagery.
\newblock In \emph{2021 International Conference on 3D Vision (3DV)}, 1075--1084.

\bibitem[{Hou et~al.(2021)Hou, Lindsay, Agu, Pedersen, Tulu, and Strong}]{HDR_bright}
Hou, S.; Lindsay, C.; Agu, E.; Pedersen, P.; Tulu, B.; and Strong, D. 2021.
\newblock HDR-Like Image Generation to Mitigate Adverse Wound Illumination Using Deep Bi-directional Retinex and Exposure Fusion.
\newblock In Papie{\.{z}}, B.~W.; Yaqub, M.; Jiao, J.; Namburete, A. I.~L.; and Noble, J.~A., eds., \emph{Medical Image Understanding and Analysis}, 307--321. Cham: Springer International Publishing.
\newblock ISBN 978-3-030-80432-9.

\bibitem[{Huang, Huang, and Wu(2024)}]{RGB2IR}
Huang, F.; Huang, W.; and Wu, X. 2024.
\newblock Enhancing Infrared Optical Flow Network Computation through RGB-IR Cross-Modal Image Generation.
\newblock \emph{Sensors}, 24(5).

\bibitem[{Huang et~al.(2018)Huang, Liu, Belongie, and Kautz}]{MUNIT}
Huang, X.; Liu, M.-Y.; Belongie, S.; and Kautz, J. 2018.
\newblock Multimodal Unsupervised Image-to-Image Translation.
\newblock In Ferrari, V.; Hebert, M.; Sminchisescu, C.; and Weiss, Y., eds., \emph{Computer Vision -- ECCV 2018}, 179--196. Cham: Springer International Publishing.
\newblock ISBN 978-3-030-01219-9.

\bibitem[{Isola et~al.(2017)Isola, Zhu, Zhou, and Efros}]{Pix2Pix}
Isola, P.; Zhu, J.-Y.; Zhou, T.; and Efros, A.~A. 2017.
\newblock Image-to-Image Translation with Conditional Adversarial Networks.
\newblock In \emph{2017 IEEE Conference on Computer Vision and Pattern Recognition (CVPR)}, 5967--5976.

\bibitem[{Johnson, Karpathy, and Fei-Fei(2016)}]{caption_relatedwork1}
Johnson, J.; Karpathy, A.; and Fei-Fei, L. 2016.
\newblock DenseCap: Fully Convolutional Localization Networks for Dense Captioning.
\newblock In \emph{2016 IEEE Conference on Computer Vision and Pattern Recognition (CVPR)}, 4565--4574.

\bibitem[{Kou et~al.(2023)Kou, Wang, Peng, Zhao, Chen, Han, Huang, Yu, and Fu}]{Infrared1}
Kou, R.; Wang, C.; Peng, Z.; Zhao, Z.; Chen, Y.; Han, J.; Huang, F.; Yu, Y.; and Fu, Q. 2023.
\newblock Infrared small target segmentation networks: A survey.
\newblock \emph{Pattern Recognition}, 143: 109788.

\bibitem[{Lee et~al.(2023)Lee, Jeon, Cho, and Kim}]{sRGB-TIR}
Lee, D.-G.; Jeon, M.-H.; Cho, Y.; and Kim, A. 2023.
\newblock Edge-guided Multi-domain RGB-to-TIR image Translation for Training Vision Tasks with Challenging Labels.
\newblock In \emph{2023 IEEE International Conference on Robotics and Automation (ICRA)}, 8291--8298.

\bibitem[{Li et~al.(2023)Li, Xu, Wu, Lu, and Kittler}]{Infrared3}
Li, H.; Xu, T.; Wu, X.-J.; Lu, J.; and Kittler, J. 2023.
\newblock LRRNet: A Novel Representation Learning Guided Fusion Network for Infrared and Visible Images.
\newblock \emph{IEEE Transactions on Pattern Analysis and Machine Intelligence}, 45(9): 11040--11052.

\bibitem[{Li, Jiang, and Han(2019)}]{caption_relatedwork4}
Li, X.; Jiang, S.; and Han, J. 2019.
\newblock Learning Object Context for Dense Captioning.
\newblock In \emph{Proceedings of the AAAI Conference on Artificial Intelligence}, volume~33, 8650--8657.

\bibitem[{Li et~al.(2024)Li, Wen, Hu, Yuan, and Zhu}]{vision_language_model2}
Li, X.; Wen, C.; Hu, Y.; Yuan, Z.; and Zhu, X.~X. 2024.
\newblock Vision-Language Models in Remote Sensing: Current progress and future trends.
\newblock \emph{IEEE Geoscience and Remote Sensing Magazine}, 12(2): 32--66.

\bibitem[{Ma et~al.(2024)Ma, Xian, Li, Li, and Zhang}]{IR-GAN}
Ma, D.; Xian, Y.; Li, B.; Li, S.; and Zhang, D. 2024.
\newblock Visible-to-infrared image translation based on an improved CGAN.
\newblock \emph{VISUAL COMPUTER}, 40(2): 1289--1298.

\bibitem[{Peng et~al.(2024{\natexlab{a}})Peng, Bashford-Rogers, Banterle, Zhao, and Debattista}]{HDRTDataset}
Peng, J.; Bashford-Rogers, T.; Banterle, F.; Zhao, H.; and Debattista, K. 2024{\natexlab{a}}.
\newblock HDRT: Infrared Capture for HDR Imaging.
\newblock \emph{arXiv preprint}, 2406.05475.

\bibitem[{Peng, Zhao, and Hu(2023)}]{DFNet}
Peng, J.; Zhao, H.; and Hu, Z. 2023.
\newblock Dynamic Fusion Network for RGBT Tracking.
\newblock \emph{IEEE Transactions on Intelligent Transportation Systems}, 24(4): 3822--3832.

\bibitem[{Peng et~al.(2023{\natexlab{a}})Peng, Zhao, Hu, Zhuang, and Wang}]{SiamIVFN}
Peng, J.; Zhao, H.; Hu, Z.; Zhuang, Y.; and Wang, B. 2023{\natexlab{a}}.
\newblock Siamese infrared and visible light fusion network for RGB-T tracking.
\newblock \emph{International Journal of Machine Learning and Cybernetics}, 14(9): 3281--3293.

\bibitem[{Peng et~al.(2023{\natexlab{b}})Peng, Zhao, Zhao, Wang, and Yao}]{CourtNet}
Peng, J.; Zhao, H.; Zhao, K.; Wang, Z.; and Yao, L. 2023{\natexlab{b}}.
\newblock CourtNet: Dynamically balance the precision and recall rates in infrared small target detection.
\newblock \emph{Expert Systems with Applications}, 233: 120996.

\bibitem[{Peng et~al.(2024{\natexlab{b}})Peng, Zhao, Zhao, Wang, and Yao}]{DBR}
Peng, J.; Zhao, H.; Zhao, K.; Wang, Z.; and Yao, L. 2024{\natexlab{b}}.
\newblock Dynamic background reconstruction via masked autoencoders for infrared small target detection.
\newblock \emph{Engineering Applications of Artificial Intelligence}, 135: 108762.

\bibitem[{Reinhard et~al.(2001)Reinhard, Adhikhmin, Gooch, and Shirley}]{ColorTransfer}
Reinhard, E.; Adhikhmin, M.; Gooch, B.; and Shirley, P. 2001.
\newblock Color transfer between images.
\newblock \emph{IEEE Computer Graphics and Applications}, 21(5): 34--41.

\bibitem[{Seetzen et~al.(2004)Seetzen, Heidrich, Stuerzlinger, Ward, Whitehead, Trentacoste, Ghosh, and Vorozcovs}]{HDR_medical}
Seetzen, H.; Heidrich, W.; Stuerzlinger, W.; Ward, G.; Whitehead, L.; Trentacoste, M.; Ghosh, A.; and Vorozcovs, A. 2004.
\newblock High dynamic range display systems.
\newblock \emph{ACM Trans. Graph.}, 23(3): 760–768.

\bibitem[{Shahzad and Jalal(2021)}]{surveillance}
Shahzad, A.~R.; and Jalal, A. 2021.
\newblock A Smart Surveillance System for Pedestrian Tracking and Counting using Template Matching.
\newblock In \emph{2021 International Conference on Robotics and Automation in Industry (ICRAI)}, 1--6.

\bibitem[{Shao et~al.(2023)Shao, Han, Debattista, and Pang}]{CapModel}
Shao, Z.; Han, J.; Debattista, K.; and Pang, Y. 2023.
\newblock Textual Context-Aware Dense Captioning With Diverse Words.
\newblock \emph{IEEE Transactions on Multimedia}, 25: 8753--8766.

\bibitem[{Shao et~al.(2022)Shao, Han, Marnerides, and Debattista}]{caption_relatedwork5}
Shao, Z.; Han, J.; Marnerides, D.; and Debattista, K. 2022.
\newblock Region-Object Relation-Aware Dense Captioning via Transformer.
\newblock \emph{IEEE Transactions on Neural Networks and Learning Systems}, 1--12.

\bibitem[{Shopovska et~al.(2023)Shopovska, Stojkovic, Aelterman, Van~Hamme, and Philips}]{HDR_ADAS}
Shopovska, I.; Stojkovic, A.; Aelterman, J.; Van~Hamme, D.; and Philips, W. 2023.
\newblock High-Dynamic-Range Tone Mapping in Intelligent Automotive Systems.
\newblock \emph{Sensors}, 23(12).

\bibitem[{Shukla et~al.(2022)Shukla, Upadhyay, Sharma, Chinnusamy, and Kumar}]{HR-NIR}
Shukla, A.; Upadhyay, A.; Sharma, M.; Chinnusamy, V.; and Kumar, S. 2022.
\newblock High-Resolution NIR Prediction from RGB Images: Application to Plant Phenotyping.
\newblock In \emph{2022 IEEE International Conference on Image Processing (ICIP)}, 4058--4062.

\bibitem[{Singh et~al.(2023)Singh, Bashford-Rogers, Marnerides, Debattista, and Hazra}]{HDR_datection}
Singh, A.~R.; Bashford-Rogers, T.; Marnerides, D.; Debattista, K.; and Hazra, S. 2023.
\newblock HDR image-based deep learning approach for automatic detection of split defects on sheet metal stamping parts.
\newblock \emph{The International Journal of Advanced Manufacturing Technology}, 125(5): 2393--2408.

\bibitem[{Tang et~al.(2023)Tang, Xiang, Zhang, Gong, and Ma}]{Infrared2}
Tang, L.; Xiang, X.; Zhang, H.; Gong, M.; and Ma, J. 2023.
\newblock DIVFusion: Darkness-free infrared and visible image fusion.
\newblock \emph{Information Fusion}, 91: 477--493.

\bibitem[{Wang et~al.(2024)Wang, Sun, Dong, and Zheng}]{PAS-GAN}
Wang, S.; Sun, G.; Dong, L.; and Zheng, B. 2024.
\newblock PAS-GAN: A GAN based on the Pyramid Across-Scale module for visible-infrared image transformation.
\newblock \emph{Infrared Physics \& Technology}, 139: 105314.

\bibitem[{Wu et~al.(2023)Wu, Hao, Wu, Xiao, He, and Yin}]{low_resolution}
Wu, H.; Hao, X.; Wu, J.; Xiao, H.; He, C.; and Yin, S. 2023.
\newblock Deep learning-based image super-resolution restoration for mobile infrared imaging system.
\newblock \emph{Infrared Physics \& Technology}, 132: 104762.

\bibitem[{Yang et~al.(2017)Yang, Tang, Yang, and Li}]{caption_relatedwork2}
Yang, L.; Tang, K.; Yang, J.; and Li, L.-J. 2017.
\newblock Dense Captioning with Joint Inference and Visual Context.
\newblock In \emph{2017 IEEE Conference on Computer Vision and Pattern Recognition (CVPR)}, 1978--1987.

\bibitem[{Yin et~al.(2019)Yin, Sheng, Liu, Yu, Wang, and Shao}]{caption_relatedwork3}
Yin, G.; Sheng, L.; Liu, B.; Yu, N.; Wang, X.; and Shao, J. 2019.
\newblock Context and Attribute Grounded Dense Captioning.
\newblock In \emph{2019 IEEE/CVF Conference on Computer Vision and Pattern Recognition (CVPR)}, 6234--6243.

\bibitem[{Zhang et~al.(2024)Zhang, Huang, Jin, and Lu}]{vision_language_model1}
Zhang, J.; Huang, J.; Jin, S.; and Lu, S. 2024.
\newblock Vision-Language Models for Vision Tasks: A Survey.
\newblock \emph{IEEE Transactions on Pattern Analysis and Machine Intelligence}, 46(8): 5625--5644.

\bibitem[{Zhang et~al.(2021)Zhang, Xiao, Wang, Peng, Chen, Zhang, Zhang, Guo, Wang, Luo, Zhou, and Xu}]{medical_diagnostics}
Zhang, L.; Xiao, M.; Wang, Y.; Peng, S.; Chen, Y.; Zhang, D.; Zhang, D.; Guo, Y.; Wang, X.; Luo, H.; Zhou, Q.; and Xu, Y. 2021.
\newblock Fast Screening and Primary Diagnosis of COVID-19 by ATR–FT-IR.
\newblock \emph{Analytical Chemistry}, 93(4): 2191--2199.
\newblock PMID: 33427452.

\bibitem[{Zhu et~al.(2017)Zhu, Park, Isola, and Efros}]{CycleGAN}
Zhu, J.-Y.; Park, T.; Isola, P.; and Efros, A.~A. 2017.
\newblock Unpaired Image-to-Image Translation Using Cycle-Consistent Adversarial Networks.
\newblock In \emph{2017 IEEE International Conference on Computer Vision (ICCV)}, 2242--2251.

\bibitem[{Özkanoğlu and Ozer(2022)}]{InfraGAN}
Özkanoğlu, M.~A.; and Ozer, S. 2022.
\newblock InfraGAN: A GAN architecture to transfer visible images to infrared domain.
\newblock \emph{Pattern Recognition Letters}, 155: 69--76.

\end{thebibliography}
